\documentclass{article}

{}
{}
{}
\usepackage{amsmath} 
\usepackage{amsthm} 
\usepackage{cite} 
\usepackage{hyperref} 
\usepackage{graphicx}
\usepackage{hyperref}
\usepackage{authblk}
\usepackage[margin=1.2cm]{geometry}
\usepackage{multicol}
\usepackage{booktabs} 

\usepackage{xcolor}

\newcommand{\revised}[1]{\textcolor{black}{#1}}

\providecommand{\captionof}[2]{#1#2}

\begin{document}

\title{\textbf{Understanding Scam Trends and Rail Paths from Reddit Self-Disclosure Narratives}}

\author[1]{Yangjun Zhang}
\author[1]{Mirko Bottarelli}
\author[1]{Mark Hooper}
\author[1]{Carsten Maple}

\affil[1]{{The Alan Turing Institute, London, UK}}


\setcounter{Maxaffil}{0}
\renewcommand\Affilfont{\itshape\small}

\date{}
\maketitle

\begin{abstract}
Online scam behavior is inherently multi-stage, and the lifecycle includes temporally ordered rails and events rather than isolated signals.
Existing works analyze characteristics of scam types and rails, but they do not track scam trends across years.
Moreover, the work on the relations between rails is hampered due to the lack of open-source datasets with annotations and coverage of different scam types.
To address these gaps, we build a dataset to analyze the yearly trend of scam characteristics and rail paths using Reddit self-disclosure narratives from 2023 to 2025.
We collect 21,304 posts from scam-related subreddits with at least one rail among identity, communication, platform, and payment for trend analysis by heuristic annotation.
Then, we label 1,800 posts containing explicit or recoverable scam chains by an LLM-assisted method for scam path analysis. The method is evaluated with human annotation.
Lastly, we run a topic model on the comments of the posts to analyze the community support behavior.
The results reveal that scam processes are predominantly multi-rail.
Across years, different scam types and rail components dominate.
Different scam types vary systematically in path complexity.
Reddit support behaviors have become more detailed over time.
\revised{
This work supports synthetic scam chain data simulation and AI-related scam risk assessment, though findings may not generalise to other platforms.}

\end{abstract}

\begin{multicols}{2}

\section{Introduction}

\revised{
Online scams are deceptive schemes targeting multiple rails, making single-rail observation insufficient for scam recognition or prevention. 
Rails refer to the channels in sequential order of a scam pathway.
E.g., in payment-related scams, the rail path can be identity rail to payment rail~\cite{islam2024multichannel}.
Scam trend may change with time.
The involvement of AI-related techniques and reliance on historical patterns alone may be inadequate for effective risk management in later years.
Understanding the scam trends and cross-rail patterns can help support scam synthetic data for scam risk management.} 



Prior work has examined specific scam types or individual channels in isolation. Job scams have been studied using survey data~\cite{goyal2023warning}. However, this work does not model the cross-rail transitions between rails, such as communication channels and payment methods, that characterize real-world scam chains. Another line of work considers the cross-rail relation. \cite{blancaflor2023social} studies 87 posts with descriptive statistics on common events, scam techniques, and victim patterns from Facebook, X, and Reddit. 1,525 posts and 1,883 comments from Reddit have been studied for scam type, initial communication method, and community support~\cite{bouma2025scam}.
\revised{The Euro Retail Payments Board (ERPB) report in 2024 frames the impersonation scam lifecycle of preparation, execution, and completion.\footnote{
\url{https://www.ecb.europa.eu/}}}
However, the above research does not annotate the complete rail flow in each case, and they do not consider the missing rails, which are not reported in user self-disclosure, e.g., someone's account is taken over, but the process before the takeover is unknown. They also do not consider the cross-rail scam pattern evolution across years.
\revised{Moreover, the ERPB has flagged AI-enabled fraud as an emerging threat, including voice cloning and deepfake impersonation.
Prior work has examined Reddit discussions of deepfakes and their societal implications \cite{gamage2022deepfakes}.
Yet little is known about AI-related scam trends over time, or how AI-enabled fake methods vary across scam types.
}

Anonymous communities, e.g., Reddit, are suitable spaces for personal disclosure and getting support from strangers, which helps explain why scam-related self-disclosure can appear there in the first place~\cite{miller2020investigating}. By building a dataset from scam-related discussions on Reddit, we aim to solve the above issues and address the following research questions: (1) How can we build a time series dataset for scam trend and rail path analysis? (2) What are the three-year trends of scam types and rails? Is AI involved in scams? How can we evaluate the data quality? Does the dataset have bias compared with previous studies? (3) What is the rail flow tendency for each type? (4) What are the user behaviours attempting to find or avoid the scam in scam self-disclosure? What are the responses of other users to help the victims?

We construct a final analytical dataset of 21,304 Reddit scam-related self-disclosure posts from 2023 to 2025, each with at least one rail and an identified scam type. We further build a stratified 1,800-post subset for LLM-assisted annotation. Based on this dataset, we analyse three-year scam trends and find that shopping scams are decreasing while impersonation scams are increasing. Scams increasingly involve emails, messaging apps, and phone calls, while WhatsApp, Telegram, and TikTok become more salient and Facebook becomes less dominant over time. \revised{AI-related scam discussions nearly quadrupled over 2023–2025, with AI functioning as a type-specific tendency.
}

We then analyse rail-path characteristics. Scams are predominantly multi-rail event chains rather than single-rail incidents. Investment and employment scams exhibit comparatively broad rail structures, while blackmail/extortion and employment scams show the longest rail paths. Communication is the dominant rail in most scam types, but investment scams are more strongly centred on platform interaction. Scam-path breadth narrows over time, while communication and payment rails become more structurally prominent within those paths.

\revised{
Finally, we analyse user verification behaviour, awareness triggers, and community responses. We find that Reddit communities have evolved from offering generic caution toward providing targeted, verification-oriented guidance, functioning as an adaptive peer support infrastructure.}

\section{Literature Review}
\subsection{Scam Lifecycle in Offline Case Studies}
Earlier work studies the scam lifecycle and victim journey, mostly focused on local mechanisms, behavioural signals, or post-victimisation outcomes, rather than the full scam path.
For example, research on advance-fee scam focused on persuasion strategies; job scam studies highlighted signals such as urgency, social proof, and payment behaviour; and breach-related work examined harassment, phishing, re-victimization, and safety concerns after the incident\cite{edwards2017scamming,abramova2023anatomy}.
These works have limited cross-rail continuity with single-stage or local-stage analyses.
Later work started to treat scams as staged processes.
In pig-butchering research, early interview-based studies already showed the gradual progression of long-term social engineering, relationship building, and financial manipulation; later work then went further by explicitly breaking the scam into lifecycle stages and discussing intervention points at different stages \cite{oak2025hello}.
Meanwhile, work on cybercrime recovery and victim journey extended attention beyond the scam event itself to stages such as recognition and recovery, highlighting post-victimisation response and support needs~\cite{chen2026harm}.
These studies remained hard to scale and hard to generalize across multiple scam types.
Our work is based on the online Reddit corpus with a temporal trend, and we analyse the cross-rail pattern of multiple scam types.

\revised{
\subsection{Online Scam Narratives and Detection}
Previous work in online scam has analysed scam-related content and community discussion in social media, such as studies compiling online banking scam content and work on cryptocurrency communities that examined victimisation experiences, newcomer guidance, community norms, and scambaiting~\cite{childs2024guess}.
Reddit-based work then looked more systematically at how users identify scams, discuss scammer tactics, ask for advice, and receive reassurance and informational support from the community~\cite{bouma2025scam,oak2025victims}, and at how public perceptions of deepfakes are shaped through Reddit discussions~\cite{gamage2022deepfakes}.
Later work moved closer to scam-path reconstruction and scam detection.
The clearest example is the exploratory pig-butchering study~\cite{oak2025hello}, which combined social media materials, abuse reports, and news cases to recover scam mechanisms, rails, and payment paths.
Another line of work has applied machine learning to fraud detection, including anomaly detection in financial transactions~\cite{rodriguez2024financial} and deepfake detection benchmarks for synthetic audio and video~\cite{khalid2021fakeavceleb}; these methods detect scam at the technical level, but do not capture the temporal pattern shift and how AI is embedded within different scam types.
%
Our work fills this gap by building a rail-based dataset across scam types, analysing temporal patterns, and AI usage.}
\section{Methodology}
\label{sec:method}
\subsection{Data Collection and Filtering}

We collect the Global Anti-Scam Alliance (GASA) report from the official website from 2023 to 2025.\footnote{\url{https://gasa.org/}}
We also collect three Reddit subreddits from 2009 to 2025: `r/Scams', `r/scammers', and `r/isthisascam', selected by ranking subreddits by post volume. 

%

We then apply a four-step filtering pipeline to construct the analytical datasets.
In \textbf{Step 1}, we remove entries whose body text is `blank', `[removed]', or `[deleted]', yielding 208,033 submissions.
In \textbf{Step 2}, we restrict to posts from 2023 to 2025 and retain only confirmed scam narratives via rule-based filtering.
A submission is considered a confirmed scam if (1) its tag explicitly indicates a scam, or (2) its comments contain affirmative confirmation (e.g., `yes, this is a scam').
We exclude posts whose titles contain uncertainty indicators (e.g., `is this a scam', `not sure') or whose tags indicate uncertainty (e.g., `Question'), unless confirmed by comments.
Posts tagged as `Not a Scam' are removed.
After this step, 124,397 submissions remain.
In \textbf{Step 3}, we apply the rule-based annotation scheme (detailed in \S\ref{sec:annotation}) to label each post along four rails and assign a scam type.
We retain only submissions where at least one rail is identified (86,843 submissions), then remove posts labeled as `Other' scam type, yielding a final analytical dataset of 21,304 posts (5,442 in 2023; 9,509 in 2024; 6,353 in 2025).
This dataset supports the three-year trend analysis.
In \textbf{Step 4}, we sample a subset for LLM agent annotation.
We apply stratified sampling by year and scam type, selecting 600 posts per year with equal quotas per scam type category where possible, yielding 1,800 posts.
This subset supports the behavioural analyses, including rail path characteristics.

\textbf{Ethics.}
All data are publicly available Reddit posts and comments, collected in compliance with platform data-use policies. We report results in aggregate and avoid disclosing personally identifying details.

\subsection{Data Annotation}
\label{sec:annotation}

\textbf{Rule-based annotation.}
We annotate each post along four rails: communication rail, interaction platform rail, payment rail, and identity rail.
Rail labels are assigned using keyword matching: the first three rails follow keywords from the GASA report (e.g., `bank transfer' for the payment rail), and the identity rail uses high-frequency keywords derived from `r/IdentityTheft' (4,825 cleaned posts from 2023 to 2025).
Each post is further assigned one primary scam type from the following categories:
`blackmail or extortion scam', `charity scam', `employment scam', `fake invoice scam',
`identity theft', `impersonation scam', `investment scam', `money recovery scam',
`romance or relationship scam', `shopping scam', and `other'.

\textbf{LLM agent annotation.}
We use Claude Sonnet 4.5 to annotate the 1,800-post subset.
The LLM is asked to infer the scam type, identify rails, extract the rail sequence, and label supporting attributes, including user verification behaviour and awareness trigger.
The prompt follows a four-part structure: (1) task definition; (2) few-shot examples; (3) boundary constraints and error cases; (4) output format.
We refine the prompt using 50 annotated pilot samples to clarify boundary cases (e.g., identity theft vs.\ impersonation), then automatically annotate the remaining 1,750 held-out samples.
The LLM agent annotation extends the rule-based schema with: (1) \textit{Rail flow}, the ordered progression of rails; and (2) \textit{Supporting attributes}, including chain completeness, initial vector, awareness trigger, and user verification behaviour.
For human validation, three Amazon MTurk workers ($>$ 97\% HIT approval rate and $>$ 500 approved HITs) independently label 170 randomly sampled posts from the 1,800-post subset, with final labels determined by majority vote. Of these, 50 overlap with the prompt refinement set, and 120 are used for human--LLM agreement evaluation.
Inter-annotator agreement among the three human annotators is measured by Fleiss' $\kappa$ for scam type. Human--LLM agreement is measured by Cohen's $\kappa$ for scam type, pooled multi-label Cohen's $\kappa$ for four-rail presence, and normalized Levenshtein similarity for rail order.

\textbf{Topic modelling.} \revised{We apply BERTopic to two sub-corpora: community comments to the posts (12 topics) and AI-related scam posts paired with non-AI controls (20 topics). Both models use the all-MiniLM-L6-v2 sentence encoder.}

\subsection{Data Annotation Quality Evaluation}

\textbf{Rule-based annotation.}
As an external quality check, we compare the resulting rail and scam-type distributions with the GASA reports; the detailed gaps and their interpretation are discussed later in \S\ref{sec:discussion}.

\textbf{LLM-assisted annotation.}
We assessed annotation reliability along two dimensions: scam type and rail group. Inter-annotator agreement was measured on all 170 human-annotated posts, with three independent annotators assigned to each item. For scam type, Fleiss' \(\kappa = 0.77\) and the average pairwise agreement was \(0.80\), indicating substantial agreement. For the four-rail labels, pairwise agreement was high across all four rails (\(0.83\)–\(1.00\)), resulting in a pooled multi-label Cohen's \(\kappa = 0.59\). Moreover, rail-order agreement, measured using average pairwise normalized Levenshtein similarity, was moderate (\(0.65\)), indicating a reasonably consistent reconstruction of temporal scam sequences despite the inherent ambiguity of free-text narratives.
Human--LLM agreement was evaluated on the 120 held-out samples. Scam type agreement reached \(\kappa = 0.87\) (accuracy \(= 0.88\)), and the pooled four-rail multi-label \(\kappa\) was \(0.69\), with per-rail accuracy ranging from \(0.87\) to \(0.98\). Rail-order agreement, again measured using normalized Levenshtein similarity, was \(0.72\), indicating reasonably strong alignment between human and LLM annotations.
These results suggest that LLM annotation is reliable for scam type and rail presence, while agreement on rail ordering remains moderate, consistent with the ambiguity already observed among human annotators.



\section{Results}

\subsection{Three-year trend of scam}
We analyse the three-year trend of scams with the 21,304 posts.
Across 2023--2025, the three-year trends of scam types show that shopping scam, investment scam, and impersonation scam are the most prominent categories in the Reddit scam dataset.
Shopping scam is the leading category in 2023 (29.4\% of classified posts) and again in 2025 (25.6\%), whereas investment scam becomes the largest category in 2024 (23.8\%). Impersonation scam grows in prominence over time, rising from 15.4\% in 2023 to 21.5\% in 2025. 
Charity scams, fake invoice scams, and money recovery scams account for only a small share throughout the period, while scams like identity theft and employment scams remain consistently visible but at lower levels. 
The results are shown in Figure 1.

\begin{center}
\includegraphics[width=0.99\linewidth]{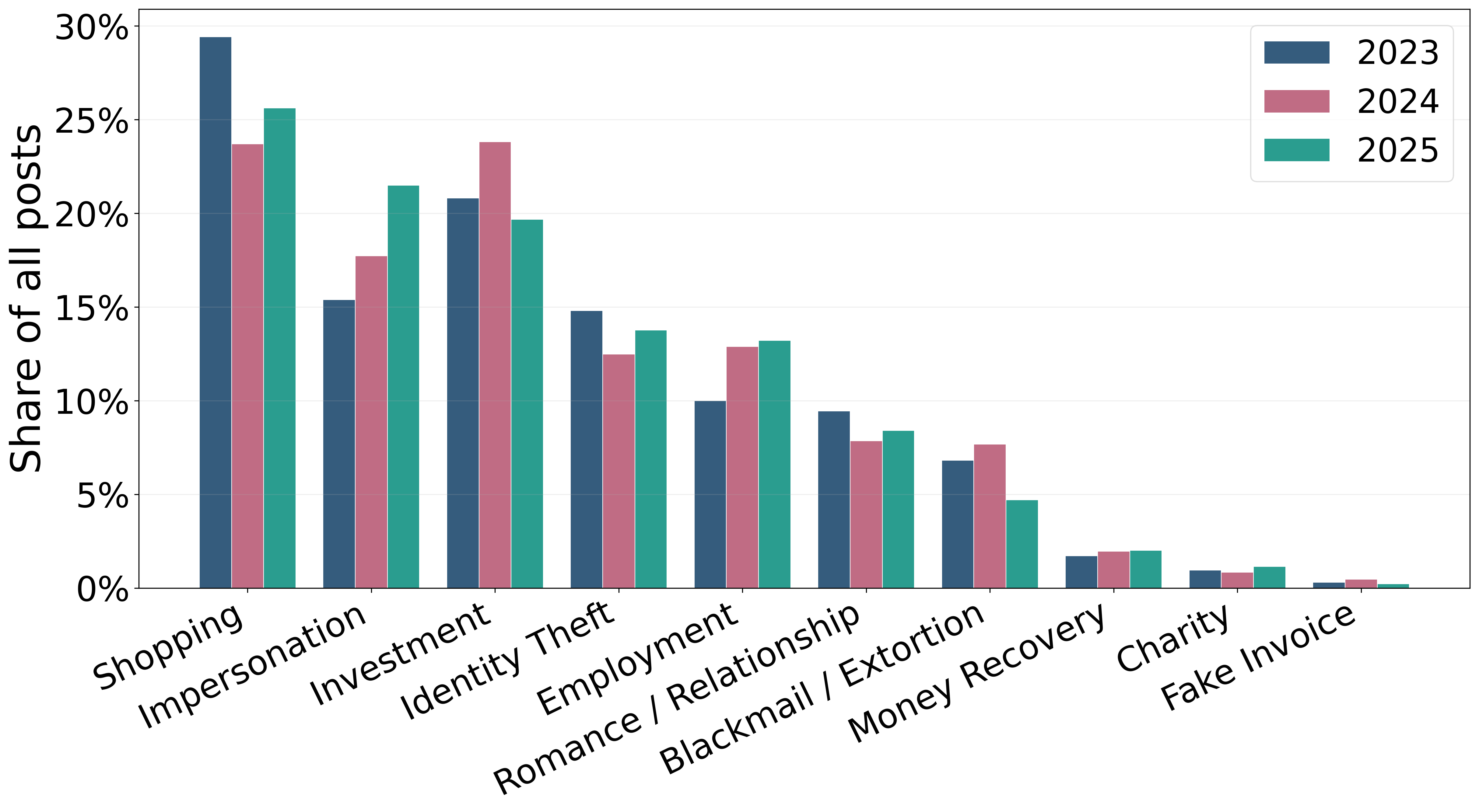}\\
\captionof{Figure 1 }{Trend of scam type.}
\label{fig:scamtype}
\end{center}

To examine the tools trend in each rail, we analyse the share of posts in the communication rail, the interaction platform rail, and the payment rail.
Overall, the communication rail is led by email and social media, while instant messaging apps and phone calls also account for a substantial share of posts. Email rises from 20.2\% of posts in 2023 to 24.2\% in 2025, and phone call increases from 8.9\% to 12.9\%.
At the platform rail, Facebook remains the most prominent platform, whereas WhatsApp, Telegram, and TikTok become increasingly salient over time.
At the payment rail, wire or bank transfer is the most common method across all three years, and cryptocurrency-related payments also increase over time.
The results are shown in Figures 2--4.


\begin{center}
\includegraphics[width=0.99\linewidth]{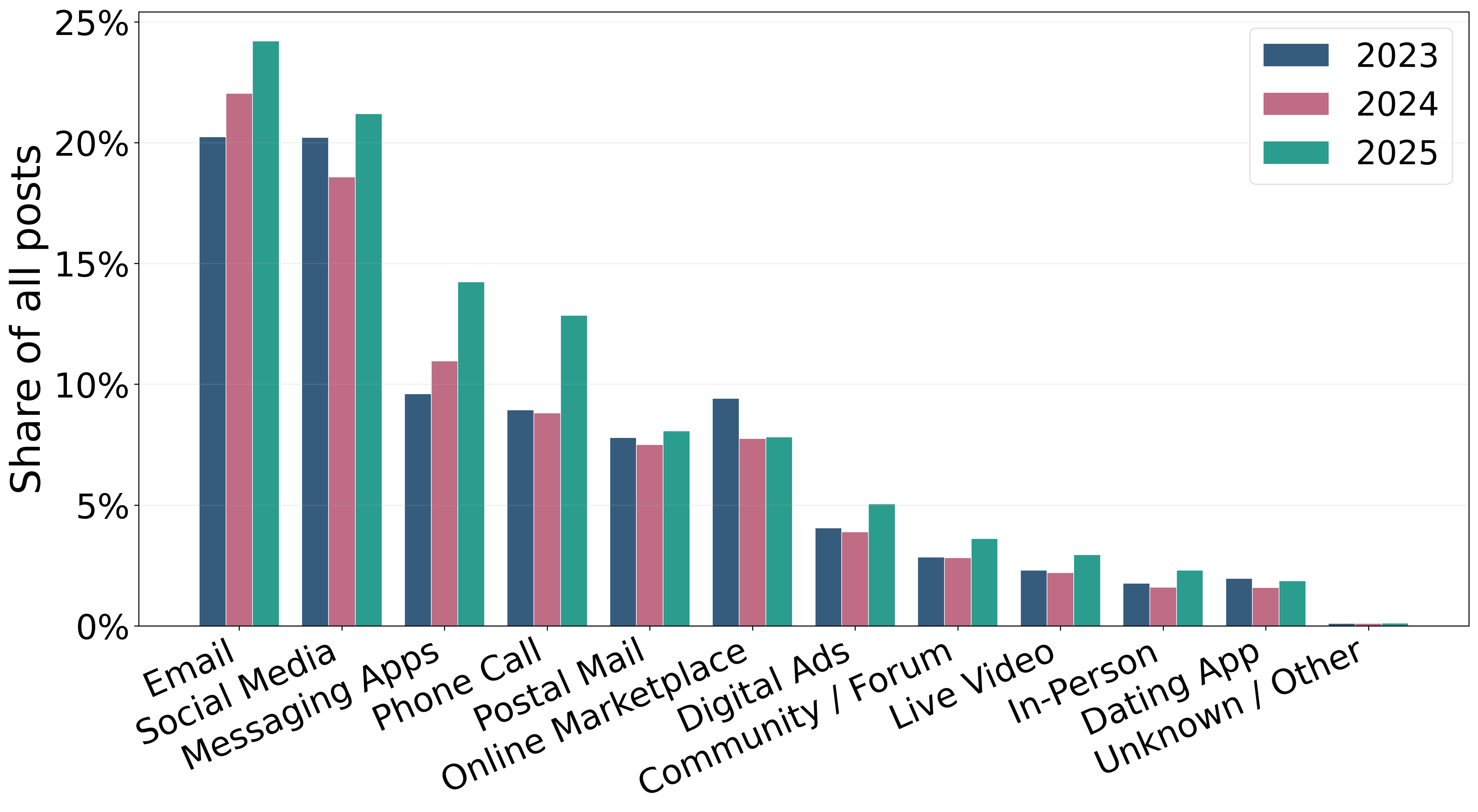}\\
\captionof{Figure 2 }{Trend of communication rail.}
\label{fig:com}
\end{center}

\begin{center}
\includegraphics[width=0.99\linewidth]{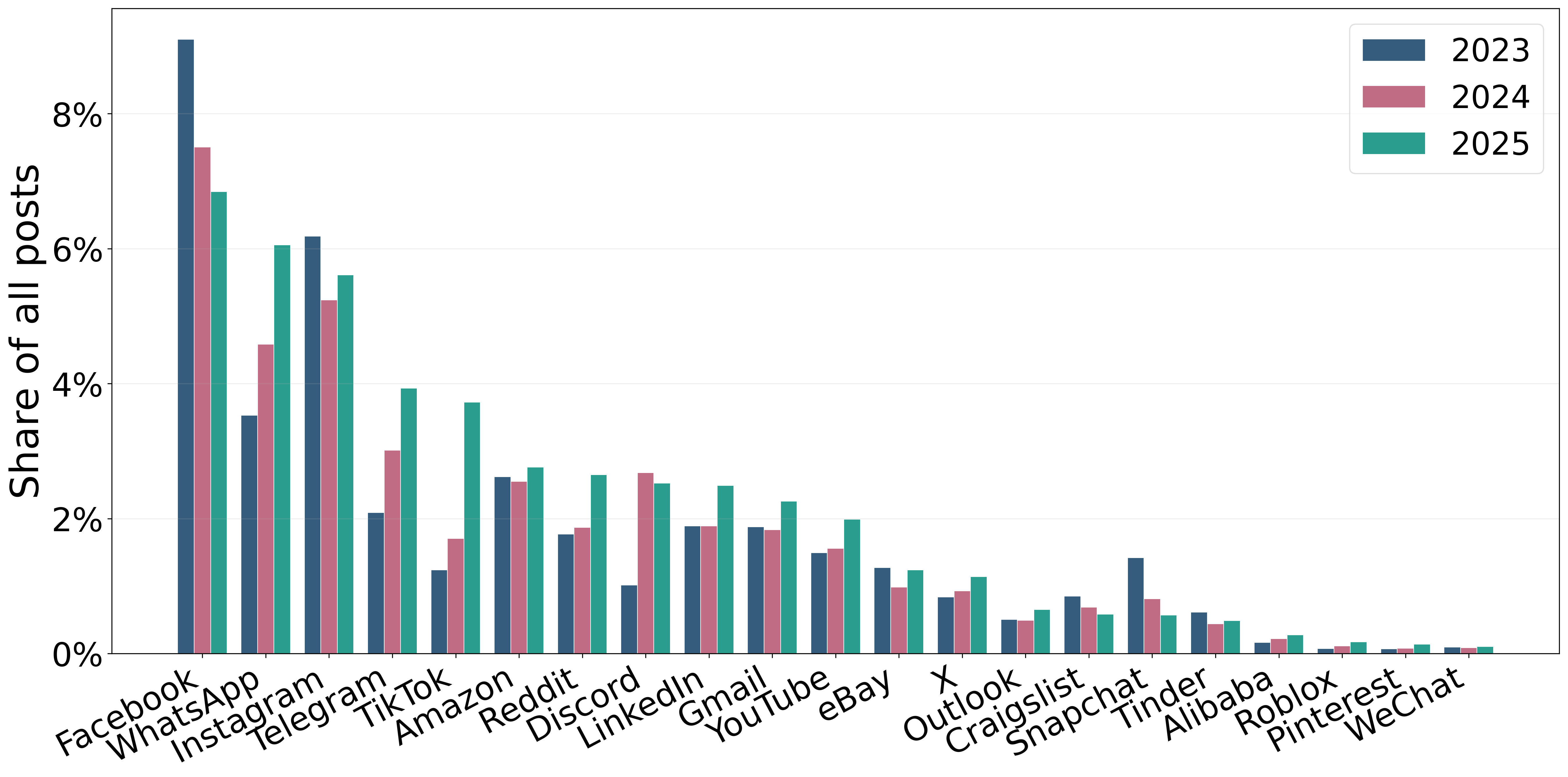}\\
\captionof{Figure 3 }{Trend of interaction platform rail.}
\label{fig:plat}
\end{center}

\begin{center}
\includegraphics[width=0.99\linewidth]{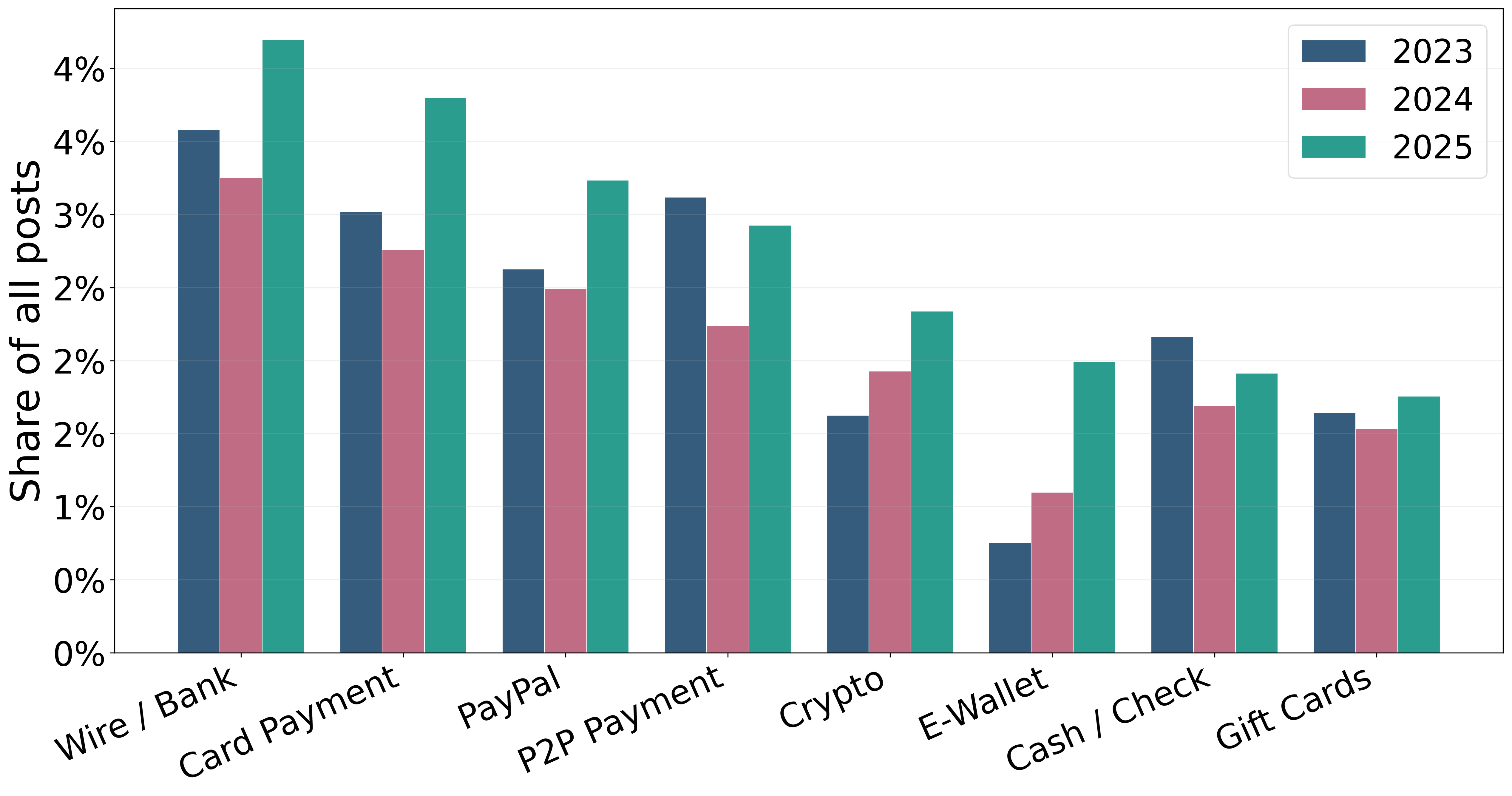}\\
\captionof{Figure 4 }{Trend of payment rail.}
\label{fig:payment}
\end{center}

%
We also examine the identity rail, which captures the types of identity-related information exploited in scams, as shown in Figure 5.
Identity-related scam posts are concentrated around traditional credentials and account-linked identifiers.
The most common identity rail is address/mail, accounting for 7.5\% of all posts, followed by email/online account (6.2\%), credit/debit card (3.3\%), bank account (1.9\%), and phone number/SIM (1.8\%). These results suggest that identity-related scam reports in the broader Reddit community are still dominated by postal misuse, account compromise, and theft of core financial identifiers.

\begin{center}
\includegraphics[width=0.99\linewidth]{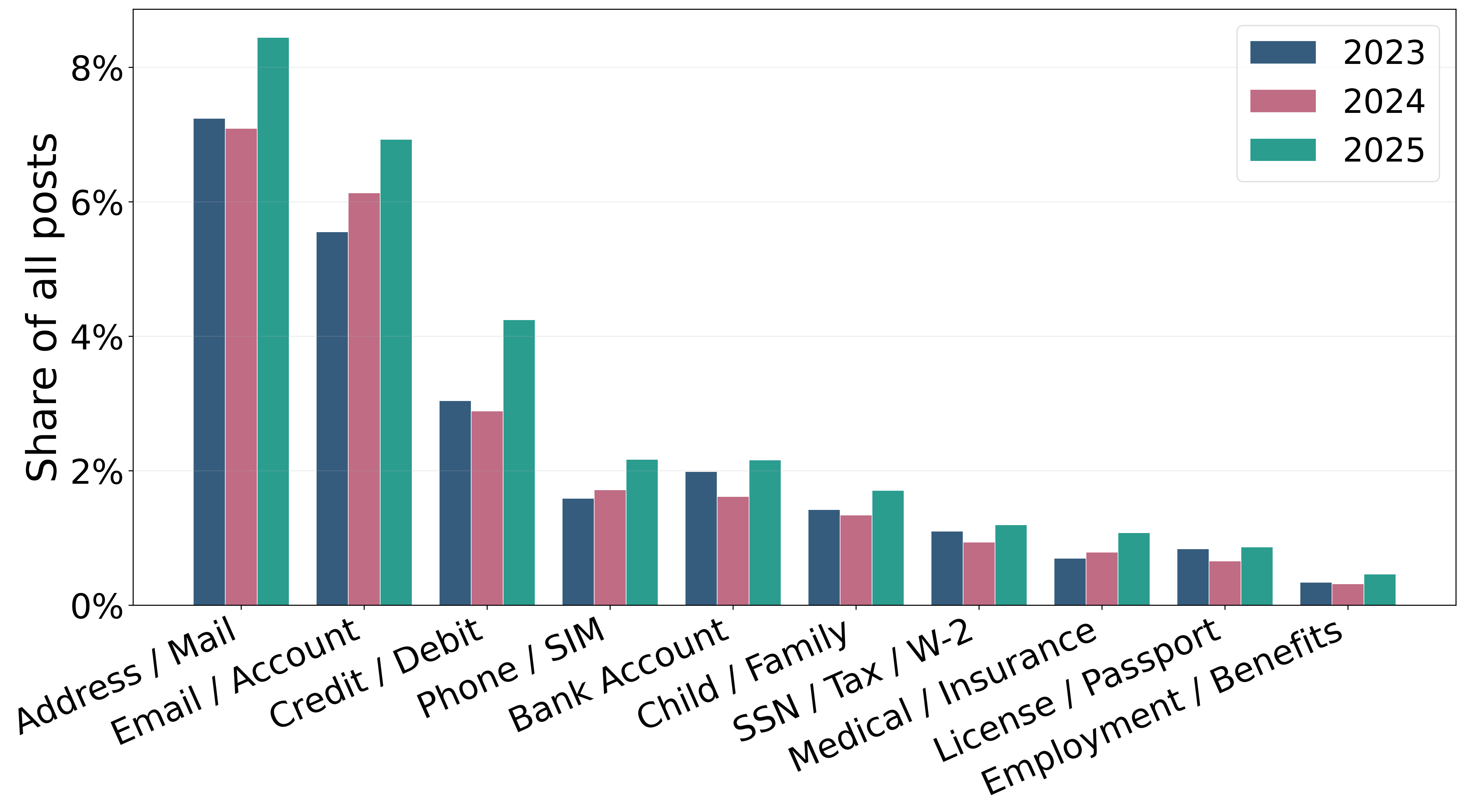}\\
\captionof{Figure 5 }{Trend of identity rail.}
\label{fig:ide}
\end{center}




\revised{
AI-related scam posts increase over time. 
Among 124,397 posts, AI-related posts grow from 7.8\textperthousand{} (2023) to 13.3\textperthousand{} (2024) and 29.7\textperthousand{} (2025); \revised{the same trend appears in the 21,304 labelled posts (9.9\textperthousand{}, 22.5\textperthousand{}, 44.1\textperthousand{}).}
AI-related discussion skews toward investment scams (36.68\% of AI-related posts vs. 20.0\% in the final dataset; 1.83 lift) and away from shopping scams (10.0\% vs. 25.9\%; 0.39 lift).
Applying BERTopic to the 548 AI-related posts in the final analytical dataset (2.57\% of the 21,304 rail-labelled posts) reveals that AI manifests as a type-specific tendency ($\chi^2=782.93$, $p<0.001$) for topics: investment scams centre on deepfake and AI-generated promotional content on platforms like YouTube; impersonation scams rely on spoofed calls, voicemail, and voice-like identity simulation; and romance scams involve fake personas on dating or social platforms.}

\subsection{Rail Path Characteristics}
We first analyse the rail composition and find that 14.3\% of cases involve a single rail while 85.7\% are multi-rail.
We treat a chain with an unknown initial rail as an incomplete chain.
In terms of rail completeness, 73.8\% of chains are complete, and 26.2\% are incomplete due to an unknown initial rail.
Among single-rail cases, 35.4\% have an unknown start path (e.g., an account is taken over but the victim cannot determine how the compromise occurred), compared to 24.7\% among multi-rail cases.

\revised{Using the four rails, we quantify scam-path complexity with rail breadth, path length, and path entropy, as shown in Table 1.
Rail breadth captures how many distinct rail dimensions appear in a post, whereas path length captures how many ordered rail steps appear in the event chain, allowing the same rail to repeat.
Entropy measures the diversity of rail types within a single scam's reported path; the scam type spanning multiple rail types evenly has high entropy.}
Under these metrics, investment scam shows the greatest cross-rail breadth (2.703), followed by employment scam (2.533) and blackmail/extortion scam (2.521).
In contrast, the fake invoice scam (2.136) is the narrowest.
The impersonation scam (2.192) and the shopping scam (2.312) also remain relatively constrained.
Path length and entropy reveal a related but not identical pattern: blackmail/extortion scam (3.116) and employment scam (3.112) have the longest paths, while employment scam also has the highest path entropy (4.690), indicating the most diverse path structures. 
At the other end, the fake invoice scam has both the shortest paths (2.270) and the lowest entropy (3.449), suggesting a more fixed and repetitive scam process.
These measures distinguish breadth from repetition.
Breadth increases when a scam spans more distinct rail dimensions, while length increases when the chain contains more sequential steps, even if some rails repeat. The money recovery scam has higher breadth than the fake invoice scam but a longer path relative to its breadth, indicating that it reuses a small set of rails across multiple steps rather than expanding into many different dimensions.

\begin{center}
\captionof{Table 1 }{Rail flow characteristics by scam type.}
\begin{tabular}{lccc}
\toprule
Type & Breadth & Length & Entropy \\
\midrule
Investment & 2.703 & 3.047 & 3.704 \\
Employment & 2.533 & 3.112 & 4.690 \\
Blackmail / extortion & 2.521 & 3.116 & 4.544 \\
Charity & 2.486 & 2.668 & 3.130 \\
Money recovery & 2.486 & 2.727 & 3.949 \\
Identity theft & 2.405 & 2.492 & 3.531 \\
Romance / relationship & 2.391 & 2.918 & 4.460 \\
Shopping & 2.312 & 2.691 & 4.065 \\
Impersonation & 2.192 & 2.641 & 4.170 \\
Fake invoice & 2.136 & 2.270 & 3.449 \\
\bottomrule
\end{tabular}
\end{center}

Degree-based analysis further shows that communication remains the dominant structural rail for several scam types, with the highest average degree per post in blackmail/extortion scam (1.737), employment scam (1.481), romance scam (1.467), and impersonation scam (1.402). However, this pattern is not universal. In investment scams, the platform has the highest average degree (1.656), exceeding payment (0.773) and communication (1.063), indicating that platform-mediated interaction plays the central organizing role in that scam type. Identity theft also differs from the broader pattern, with identity showing the highest average degree (1.132).
Consistent with this pattern, Figure 6 illustrates the rail-flow structure of the investment scam, highlighting the most common transitions among communication, platform, identity, and payment.

\begin{center}
\includegraphics[width=0.9\linewidth]{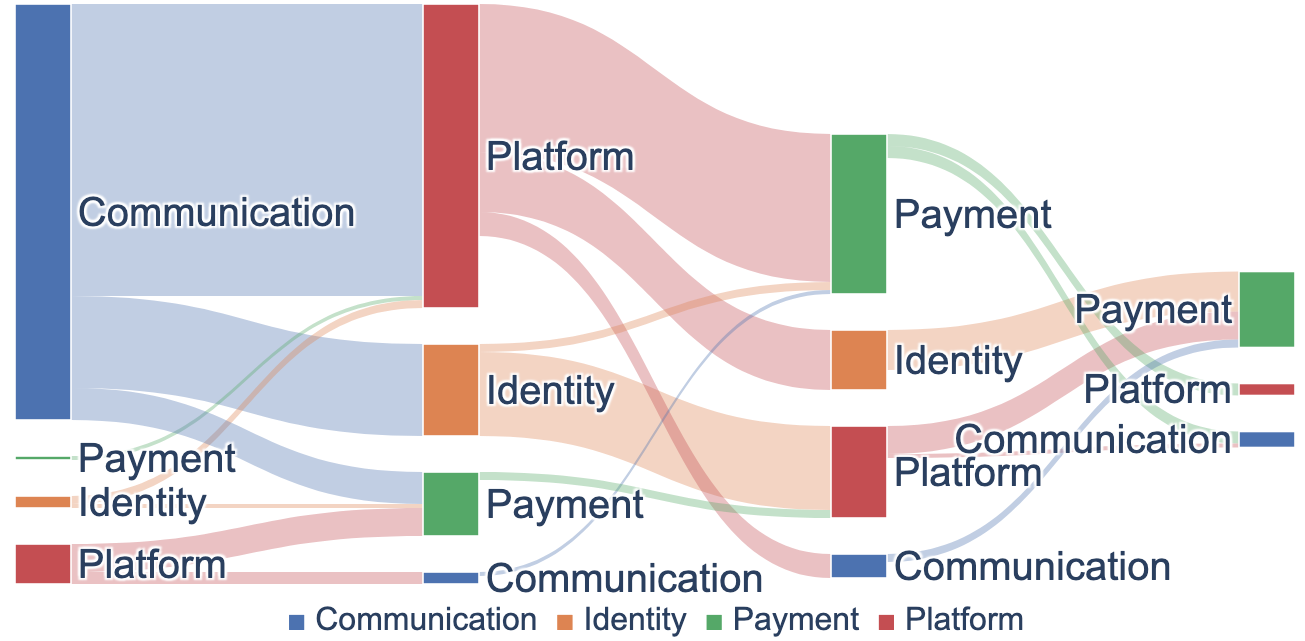}\\
\captionof{Figure 6 }{Sankey chart of investment scam showing rail flow.}
\label{fig:sankey}
\end{center}

Across years, overall scam paths do not follow a simple rebound pattern. At the aggregate level, mean rail breadth decreases from 2.64 in 2023 to 2.34 in 2024 and remains similarly low at 2.32 in 2025 ($p < 0.001$), suggesting that scam chains span fewer distinct rail dimensions over time. In contrast, mean path length changes only modestly, from 2.82 in 2023 to 2.77 in 2024 and 2.94 in 2025 ($p = 0.074$), indicating no statistically reliable temporal shift in overall path length. A different pattern appears in rail degree: communication degree rises from 1.00 in 2023 to 1.28 in 2024 and 1.59 in 2025 ($p < 0.001$), while payment degree increases from 0.44 to 0.51 and then 0.63 ($p = 0.042$). Rather than showing temporary simplification in 2024 followed by a general rebound, these results suggest that scam paths become narrower across time, while communication and payment become more structurally prominent within those paths.

Within scam types, the blackmail/extortion scam shows the clearest temporal shift. Its mean rail breadth drops from 2.99 in 2023 to 2.32 in 2024 and then to 2.17 in 2025 ($p < 0.001$), while mean path length follows a dip-and-rebound pattern, falling from 3.23 to 2.85 and then increasing to 3.27 ($p = 0.019$).
Employment scam shows a different form of temporal change: path length increases from 2.86 in 2023 to 3.01 in 2024 and 3.46 in 2025 ($p = 0.047$), while communication degree rises from 1.01 to 1.47 and then 1.92 ($p < 0.001$).
In contrast, the investment scam no longer shows a statistically reliable temporal shift in either breadth or path length under the human-preferred merged labels. Rather than indicating a uniform trend toward either simplification or complication, these results suggest that scam-path complexity changes unevenly across scam types, with some becoming narrower while others grow more communication-heavy over time.

\subsection{Scam Recognition and Community Response Analysis}
We next examine victim scam verification behaviour and scam awareness triggers. Verification strategies vary substantially across scam types.
Identity theft shows the highest rate of asking network (36.4\%), suggesting that victims often seek confirmation from others when trying to understand account compromise or identity-related misuse.
Employment scams are also verification-heavy, with the highest rate to verify entity (33.8\%) and a relatively elevated rate of checking reviews (7.8\%), consistent with users investigating employers, websites, and company reputations.
Blackmail/extortion scams show the highest rate of no verification (26.2\%), indicating that these cases more often move forward without systematic checking.
Overall, the most common verification behaviours are common sense, asking network, and verifying the entity, while more specialized behaviours, such as using tools, freezing credit, or checking records, remain rare.

Awareness triggers also differ systematically across scam types.
Fake invoice scams are overwhelmingly self-discovered (78.8\%), while identity theft has the highest rate of platform notification (20.7\%), reflecting cases where victims learn of the problem from account or system alerts rather than from the scammer directly.
Romance scams stand out for a relatively high rate of third-party informed (27.1\%), and investment scams show the highest rate of financial loss as the awareness trigger (26.6\%).
Across years, the overall distribution of awareness triggers changes significantly ($\chi^2 = 38.28$, $p < 0.001$): self-discovered rises from 54.7\% in 2023 to 61.8\% in 2025, while platform notification declines from 4.9\% to 1.2\%.
Verification behaviours are not fully stable over time at the aggregate level.
Verifying entity increases from 17.2\% in 2023 to 23.9\% in 2025 ($p = 0.020$), no verification rises from 10.7\% to 18.8\% ($p < 0.001$), and common sense declines from 55.7\% to 39.6\% ($p < 0.001$).
At the scam-type level, employment scam still shows the clearest temporal change in awareness-trigger distribution ($p = 0.030$), while identity theft also shows a significant shift over time ($p = 0.015$).

\revised{
Next, we analyse community comment topics.
The dominant topics are scam identification and general advice (26.1\%), banking and financial security guidance (17.2\%), and send-money and payment warnings (6.0\%), together covering nearly half of all replies, with smaller clusters capturing email authenticity, phone verification, empathetic support, and reporting guidance.
Across years, topic distributions shift significantly ($\chi^2 = 267.86$, $p < 0.001$). Phone and contact verification rises from 3.0\% to 6.8\%, and email/message authenticity assessment increases from 4.4\% to 9.5\%. Conversely, send-money and payment warnings decline sharply from 13.8\% to 5.2\%, and banking guidance falls modestly from 27.4\% to 25.9\%. 
This pattern reflects a shift from broad heuristic warnings toward more concrete, verification-oriented support.}

\section{Discussion and Limitations}
\label{sec:discussion}
We compare the trends and patterns with the GASA survey.
Because the GASA scam-type breakdown is not fully consistent across years, we do not use it to analyse cross-year trends.
The set of comparable scam-type categories changes across reports (e.g., 9 comparable categories in 2023 versus 10 in 2024 and 2025), and some categories are missing or defined differently across years.
We therefore restrict direct percentage-gap comparisons to 2025, which provides the closest match to the taxonomy used in our final benchmark alignment.
Using a 2025 taxonomy-aligned comparison and standardizing both sources over the 10 shared scam-type categories, Reddit over-represents shopping scams (+10.0 pp), impersonation scams (+9.4 pp), and investment scams (+6.0 pp) relative to GASA. By contrast, fake invoice scams (-9.6 pp), charity scams (-8.8 pp), and money recovery scams (-7.5 pp) are under-represented.
Overall agreement is moderate (\(JSD = 0.35\), \(\rho = 0.64\)), suggesting that the Reddit corpus reflects a related but platform-skewed scam profile rather than a population-representative prevalence distribution.
%
\revised{
We further compare our findings with the ERPB reports on Fraud.
The two sources align on three aspects: investment and impersonation scams as top prevalent fraud types; a multi-phase fraud structure mapping to our multi-rail chain model; and voice cloning and deepfake video as emerging AI-enabled threats, corroborated by the rapid growth of AI-related discussions in our dataset.}

To scale annotation, we rely on rule-based and LLM-assisted annotation.
This may introduce bias to the scam type distribution, although we have compared the distribution with GASA.
While the scam type and rail presence agreement are strong, rail ordering remains moderately reliable.
\revised{Chain order results should be interpreted with this uncertainty in mind, and we have not included causal inference for chains. The chains are developed via temporal order.}
The data is also limited by its self-reported nature.
\revised{All claims are scoped to the Reddit corpus.
Generalising the findings in this paper would require collaboration with additional data sources.}

\section{Conclusion and Future Work}

In summary, this work builds a Reddit-based scam self-disclosure dataset.
We further analyse scam prevalence and three-year trends.
We also delve into the chain structure of scam processes.
The analysis results indicate that most scams are multi-rail.
From 2023 to 2025, impersonation shows an increasing trend; email is used more for scammer communication; apps like Facebook and Instagram are less involved, while WhatsApp, Telegram, and TikTok are increasing.
Communication is the dominant rail, and some types of scams are more complex in path structure, e.g., investment and employment, while some are simpler, e.g., fake invoice.
In addition, the Reddit community provides concrete support for scam victims.
\revised
{
In the future, we will work on two directions: (i) extending the dataset with external sources such as law enforcement records and financial institution reports to improve generalizability, and (ii) moving from analysis to causal inference on scam paths.}

\bibliographystyle{unsrt}
\bibliography{reference}

\end{multicols}




\end{document}